\documentclass{article}

\usepackage[nonatbib,final]{neurips_2019}

\usepackage[utf8]{inputenc} 
\usepackage[T1]{fontenc}    
\usepackage{hyperref}       
\usepackage{url}            
\usepackage{booktabs}       
\usepackage{amsfonts}       
\usepackage{nicefrac}       
\usepackage{microtype}      

\usepackage[usenames,dvipsnames]{xcolor}
\definecolor{purple}{rgb}{0.3,0.0,.4}

\definecolor{darkgreen}{RGB}{0,120,0}

\title{Factor Group-Sparse Regularization for Efficient Low-Rank Matrix Recovery}

\author{%
  Jicong Fan\\
  Cornell University\\
  Ithaca, NY 14850 \\
  \texttt{jf577@cornell.edu} \\
  \And
   Lijun Ding \\
  Cornell University\\
  Ithaca, NY 14850 \\
   \texttt{ld446@cornell.edu} \\
   \And
   Yudong Chen \\
  Cornell University\\
  Ithaca, NY 14850 \\
   \texttt{yudong.chen@cornell.edu} \\
   \And
   Madeleine Udell \\
  Cornell University\\
  Ithaca, NY 14850 \\
   \texttt{udell@cornell.edu} \\
}
\usepackage{graphicx}
\usepackage{algorithm}
\usepackage{algorithmicx}
\usepackage{algpseudocode}
\usepackage{mathrsfs}
\usepackage{amsthm,amsmath,amssymb}

\usepackage{cases}
\usepackage{bm}
\usepackage{caption2}

\newtheorem{theorem}{Theorem}

\usepackage{enumitem}
\usepackage{bbm}

\begin{document}
\maketitle

\begin{abstract}
This paper develops a new class of nonconvex regularizers for low-rank matrix recovery.
Many regularizers are motivated as convex relaxations of the \emph{matrix rank} function.
Our new factor group-sparse regularizers are motivated as a
relaxation of the \emph{number of nonzero columns} in a factorization of the matrix.
These nonconvex regularizers are sharper than the nuclear norm;
indeed, we show they are related to Schatten-$p$ norms with arbitrarily small $0 < p \leq 1$.
Moreover, these factor group-sparse regularizers
can be written in a factored form
that enables efficient and effective nonconvex optimization;
notably, the method does not use singular value decomposition.
We provide generalization error bounds for low-rank matrix completion
which show improved upper bounds for Schatten-$p$ norm reglarization as $p$ decreases.
Compared to the max norm and the factored formulation of the nuclear norm,
factor group-sparse regularizers are more efficient, accurate,
and robust to the initial guess of rank.
Experiments show promising performance of factor group-sparse regularization
for low-rank matrix completion and robust principal component analysis.
\end{abstract}

\section{Introduction}
Low-rank matrices appear throughout the sciences and engineering,
in fields as diverse as computer science, biology, and economics
\cite{udell2019big}.
One canonical low-rank matrix recovery problem is
low-rank matrix completion (LRMC) \cite{CandesRecht2009,toh2010accelerated,MC_Recht2011:SAM,foygel2011concentration,hardt2014understanding,
MC_icml2014c1_chenc14,shamir2014matrix,sun2016guaranteed,Fan_2019_CVPR}, which aims to recover a low-rank matrix from a few entries.
LRMC has been used to impute missing data,
make recommendations, discover latent structure,
perform image inpainting, and classification \cite{NIPS2010_3932,udell2016generalized,udell2019big}.
Another important low-rank recovery problem
is robust principal components analysis (RPCA) \cite{RPCA,feng2013online,zhao2014robust,pimentel2017random,rkpca_fan}, which aims to recover a low-rank matrix from sparse but arbitrary corruptions. RPCA is often used for denoising and image/video processing \cite{8425659}.

\paragraph{LRMC} Take LRMC as an example.
Suppose $\bm{M}\in\mathbb{R}^{m\times n}$ is a low-rank matrix with $\textup{rank}(\bm{M})=r\ll \min(m,n)$.
We wish to recover $\bm{M}$ from a few observed entries.
Let $\Omega\subset [m]\times [n]$ index the observed entries.
Suppose $\textup{card}(\Omega)$, the number of observations, is sufficiently large.
A natural idea is to recover the missing entries by solving
\begin{equation}\label{opt: rankminLRMC}
\mathop\textup{minimize}_{\bm{X}}\ \textup{rank}(\bm{X}),\ \textup{subject to}\ \mathcal{P}_\Omega(\bm{X})=\mathcal{P}_\Omega(\bm{M}),
\end{equation}
where the operator $\mathcal{P}_\Omega:\mathbb{R}^{m\times n}\rightarrow \mathbb{R}^{m\times n}$ acts on any $\bm{X}\in\mathbb{R}^{m\times n}$ in the following way: $( \mathcal{P}_\Omega(\bm{X}) )_{ij}=\bm{X}_{ij}\ \textup{if}\ (i,j)\in\Omega\ \textup{and}\ 0\ \textup{if}\ (i,j)\notin\Omega$.
However, since the direct rank minimization problem \eqref{opt: rankminLRMC}
is NP-hard, a standard approach is to replace the rank with a tractable surrogae $R(\bm{X})$ and solve
\begin{equation}\label{opt: rankminLRMC_regu}
\mathop\textup{minimize}_{\bm{X}}\ R(\bm{X}),\ \textup{subject to}\ \mathcal{P}_\Omega(\bm{X})=\mathcal{P}_\Omega(\bm{M}).
\end{equation}
Below we review typical choices of $R(\bm{X})$ to provide context for our factor group-sparse regularizers.

\paragraph{Nuclear and Schatten-$p$ norms}
One popular convex surrogate function for the rank function is the nuclear norm (also called trace norm), which is defined as the sum of singular values:
\vspace{-2pt}
\begin{equation}
\Vert \bm{X}\Vert_\ast:=\sum_{i=1}^{\min(m,n)}\sigma_i(\bm{X}),
\end{equation}
where $\sigma_i(\bm{X})$ denotes the $i$-th largest singular value of $\bm{X}\in\mathbb{R}^{m\times n}$.
Variants of the nuclear norm,
including the \emph{truncated nuclear norm} \cite{MC_TNNR_PAMI2013} and \emph{weighted nuclear norm} \cite{gu2014weighted},
sometimes perform better empirically on imputation tasks.

The Schatten-$p$ norms\footnote{Note that formally $\Vert \cdot\Vert_{\textup{S}p}$ with $0\leq p<1$ is a quasi-norm, not a norm; abusively, we still use the term ``norm'' in this paper.}
with $0\leq p\leq 1$ \cite{MC_ShattenP_AAAI125165,Liu2014218,pmlr-v70-ongie17a}
form another important class of rank surrogates:
\begin{equation}
\Vert \bm{X}\Vert_{\textup{S}p}:=\Big(\sum_{i=1}^{\min(m,n)}\sigma_i^p(\bm{X})\Big)^{1/p}.
\end{equation}
For $p=1$, we have $\Vert \bm{X}\Vert_{\textup{S}_1}^1=\Vert \bm{X}\Vert_\ast$, the nuclear norm.
For $0<p<1$, $\Vert \bm{X}\Vert_{\textup{S}_p}^p$ is a nonconvex surrogate for $\textup{rank}(\bm{X})$.
In the extreme case $p=0$, $\Vert \bm{X}\Vert_{\textup{S}_0}^0=\textup{rank}(\bm{X})$,
which is exactly what we wish to minimize.
Thus we see $\Vert \bm{X}\Vert_{\textup{S}_p}^p$ with $0<p<1$ interpolates between the rank function and the nuclear norm. Instead of (\ref{opt: rankminLRMC}), we hope to solve
\begin{equation}\label{opt: SpminLRMC}
\mathop\textup{minimize}_{\bm{X}} \Vert \bm{X}\Vert_{\textup{S}_p}^p,\ \textup{subject to}\ \mathcal{P}_\Omega(\bm{X})=\mathcal{P}_\Omega(\bm{M}),
\end{equation}
with $0<p\leq 1$. Smaller values of $p$ ($0<p\leq 1$)
are better approximations of the rank function
and may lead to better recovery performance for LRMC and RPCA.
However, for $0<p<1$ the problem~(\ref{opt: SpminLRMC}) is nonconvex,
and it is not generally possible to guarantee
we find a global optimal solution.
Even worse, common algorithms for minimizing the nuclear norm and
Schatten-$p$ norm cannot scale to large matrices because they compute
the singular value decomposition (SVD) in every iteration of the optimization
\cite{CandesRecht2009, toh2010accelerated, mohan2012iterative}.

\paragraph{Factor regularizations}
A few SVD-free methods have been develoepd
to recover large low-rank matrices. For example, the work in \cite{srebro2005maximum,rennie2005fast} uses the well-known fact that
\begin{equation}\label{Eq.MF_NN_0}
\Vert \bm{X}\Vert_\ast=\min_{\bm{A}\bm{B}=\bm{X}}\Vert \bm{A}\Vert_F\Vert \bm{B}\Vert_F
=\min_{\bm{A}\bm{B}=\bm{X}}\dfrac{1}{2}\big(\Vert \bm{A}\Vert_F^2+\Vert \bm{B}\Vert_F^2\big),
\end{equation}
where $\bm{A}\in\mathbb{R}^{m\times d}$, $\bm{B}\in\mathbb{R}^{d\times n}$, and $d\geq \textup{rank}(\bm{X})$. For LRMC they suggest solving
\begin{equation}\label{Eq.MF_r}
\mathop\textup{minimize}_{\bm{A},\bm{B}} \dfrac{1}{2}\Vert \mathcal{P}_\Omega(\bm{M}-\bm{A}\bm{B})\Vert_F^2+\dfrac{\lambda}{2}\big(\Vert \bm{A}\Vert_F^2+\Vert \bm{B}\Vert_F^2\big).
\end{equation}
In this paper, we use the name \emph{factored nuclear norm} (F-nuclear norm for short)
for the variational characterization of nuclear norm as $\min_{\bm{A}\bm{B}=\bm{X}}\tfrac{1}{2}\big(\Vert \bm{A}\Vert_F^2+\Vert \bm{B}\Vert_F^2\big)$ in \eqref{Eq.MF_NN_0}.
This expression matches the nuclear norm when $d$ is chosen large enough.
Srebro and Salakhutdinov \cite{NIPS2010_4102} proposed a weighted F-nuclear norm;
the corresponding formulation of matrix completion is similar to (\ref{Eq.MF_r}).
Note that to solve (\ref{Eq.MF_r}) we must first choose the value of $d$.
We require $d\geq r:=\textup{rank}(\bm{M})$ to be able to recover (or even represent) $\bm{M}$.
Any $d\geq r$ gives the same solution $\bm{A}\bm{B}$ to (\ref{Eq.MF_r}).
However, as $d$ increases from $r$,
the difficulty of optimizing the objective increases.
Indeed, we observe in our experiments that the recovery error is larger for large $d$ using standard algorithms,
particularly when the proportion of observed entries is low.
In practice, it is difficult to guess $r$, and generally a very large $d$ is required.
The methods of \cite{MC_MF_Wen2012} and \cite{tan2014riemannian} estimate $r$ dynamically.

Another SVD-free surrogate of rank is the max norm, proposed by Srebro and Shraibman \cite{srebro2005rank}:
\begin{equation}\label{Eq.maxnorm}
\begin{aligned}
\Vert \bm{X}\Vert_{\textup{max}}=&\min_{\bm{A}\bm{B}=\bm{X}}\big(\max_i\Vert\bm{a}_{i}\Vert\big)\big(\max_j\Vert\bm{b}_{j}\Vert\big),
\end{aligned}
\end{equation}
where $\bm{a}_{i}$ and $\bm{b}_{j}$ denotes the $i$-th row of $\bm{A}$ and the $j$-th row of $\bm{B}^T$ respectively.
Lee et al.\ \cite{lee2010practical} proposed several efficient algorithms to solve optimization problems with the max norm.
Foygel and Srebro \cite{foygel2011concentration} provided
recovery guarantees for LRMC using the max norm as a regularizer.

Another very different approach uses implicit regularization.
Gunasekar et al.\ \cite{gunasekar2017implicit} show that for full dimensional factorization
without any regularization, gradient descent with small enough step size and initialized close enough to the origin
converges to the minimum nuclear norm solution.
However, convergence slows as the initial point and step size
converge to zero, making this method impractical.

Shang et al. \cite{shang2016tractable} provided the following characterization of the Schatten-1/2 norm:
\begin{equation}\label{Eq.2nuclearnorm}
\Vert\bm{X}\Vert_{\textup{S}_{1/2}}=\min\limits_{\bm{A}\bm{B}=\bm{X}}\Vert\bm{A}\Vert_\ast\Vert\bm{B}\Vert_\ast=\min\limits_{\bm{A}\bm{B}=\bm{X}}\big(\tfrac{\Vert\bm{A}\Vert_\ast+\Vert\bm{B}\Vert_\ast}{2}\big)^2.
\end{equation}
Hence instead of directly minimizing $\Vert\bm{X}\Vert_{\textup{S}_{1/2}}^{1/2}$,
one can minimize $\Vert\bm{A}\Vert_\ast+\Vert\bm{B}\Vert_\ast$,
which is much easier when $r\leq d\ll \min(m,n)$.
But again, this method and its extension $\Vert\bm{A}\Vert_\ast+\tfrac{1}{2}\Vert\bm{B}\Vert_F^2$ proposed in \cite{shang2017bilinear} require $d \geq r$, and the computational cost increases with larger $d$.
Figure \ref{Fig_MC_clean_curve}(d) shows these approaches are nearly as expensive as
directly minimizing $\Vert\bm{X}\Vert_{\textup{S}_p}^{p}$ when $d$ is large.
We call the regularizers $\min_{\bm{A}\bm{B}=\bm{X}} (\Vert\bm{A}\Vert_\ast+\Vert\bm{B}\Vert_\ast )$ and $\min_{\bm{A}\bm{B}=\bm{X}} (\Vert\bm{A}\Vert_\ast+\tfrac{1}{2}\Vert\bm{B}\Vert_F^2)$ the \emph{Bi-nuclear norm} and \emph{F$^2+$nuclear norm} respectively.

\paragraph{Our methods and contributions}
In this paper, we propose a new class of factor group-sparse regularizers (FGSR)
as a surrogate for the rank of $\bm{X}$.
To derive our regularizers, we introduce the factorization $\bm{A}\bm{B}=\bm{X}$ and seek to minimize the number of nonzero columns of $\bm{A}$ or $\bm{B}^T$. Each factor group-sparse regularizer is formed by taking the convex relaxation of the number of nonzero columns. These regularizers
are convex 
functions of the factors $\bm{A}$ and $\bm{B}$ but capture the nonconvex Schatten-$p$ (quasi-)norms of $\bm{X}$
using the nonconvex factorization constraint $\bm{X} = \bm{A}\bm{B}$.
\begin{itemize}[leftmargin=20pt,itemsep=0pt]
\item We show that these regularizers match arbitrarily sharp Schatten-$p$ norms: for each $0 <p'\leq 1$,
there is some $p < p'$ for which we exhibit a factor group-sparse regularizer equal to the sum
of the $p$th powers of the singular values of $\bm{X}$.
\item For a class of $p$, we propose a generalized factorization model that enables us to minimize $\Vert\bm{X}\Vert_{\textup{S}_{p}}^{p}$ without performing the SVD.
\item We show in experiments that the resulting algorithms improve on state-of-the-art methods for LRMC and RPCA.
\item We prove generalization error bounds for LRMC with Schatten-$p$ norm regularization, which explain the superiority of our methods over nuclear norm minimization.
\end{itemize}

\paragraph{Notation} Throughout this paper, $\Vert\cdot\Vert$ denotes the Euclidean norm of a vector argument.
We factor $\bm{X} \in \mathbb{R}^{m\times n}$ as $\bm{A}=[\bm{a}_1,\bm{a}_2,\cdots,\bm{a}_d] \in\mathbb{R}^{m\times d}$ and $\bm{B}=[\bm{b}_1,\bm{b}_2,\cdots,\bm{b}_d]^T \in\mathbb{R}^{d\times n}$,
where $d \geq r := \textup{rank}(\bm{X})$, and $\bm{a}_j$ and $\bm{b}_j$ are column vectors.
Without loss of generality, we assume $m\leq n$.
All proofs appear in the supplement.

\section{FGSRs match Schatten-$p$ norms with $p=\frac{2}{3}$ or $\frac{1}{2}$.}
Let $\textup{nnzc}(\bm{A})$ denote the number of nonzero columns of matrix $\bm{A}$.
Write the rank of $\bm{X}\in\mathbb{R}^{m\times n}$ as
\begin{equation}
  \label{eq-rank-as-nnzc}
\textup{rank}(\bm{X})=\min_{\bm{A}\bm{B}=\bm{X}}\textup{nnzc}(\bm{A})=\min_{\bm{A}\bm{B}=\bm{X}}\textup{nnzc}(\bm{B}^T)
=\min_{\bm{A}\bm{B}=\bm{X}}\dfrac{1}{2}\big(\textup{nnzc}(\bm{A})+\textup{nnzc}(\bm{B}^T)\big).
\end{equation}
Now relax: notice $\textup{nnzc}(\bm{A}) 
\geq \sum_{j=1}^d \|\bm{a_j}\|$
when $\|\bm{a_j}\| \leq 1$ for each column $j$.
We show that using this relaxation in (\ref{eq-rank-as-nnzc}) gives
a factored characterization of the Schatten-$p$ norm with $p=\frac{1}{2}$ or $\frac{2}{3}$.
\begin{theorem}\label{the_1}
Fix $\alpha>0$. For any matrix $\bm{X}\in\mathbb{R}^{m\times n}$ with $\textup{rank}(\bm{X})=r\leq d\leq\min(m,n)$,
\begin{align}
\min\limits_{\bm{X} = \sum_{j=1}^d\bm{a}_j\bm{b}_j^T}\sum_{j=1}^d\Vert \bm{a}_{j}\Vert+\Vert \bm{b}_{j}\Vert
=& 2\sum_{j=1}^r\sigma_j^{1/2}(\bm{X}) \label{the_1a} \\
\min\limits_{\bm{X} = \sum_{j=1}^d\bm{a}_j\bm{b}_j^T}\sum_{j=1}^d\Vert \bm{a}_{j}\Vert+\dfrac{\alpha}{2}\Vert \bm{b}_{j}\Vert^2
=& \dfrac{3\alpha^{1/3}}{2}\sum_{j=1}^r\sigma_j^{2/3}(\bm{X}). \label{the_1b}
\end{align}
\end{theorem}
Denote the SVD of $\bm{X}$ as $\bm{X}=\bm{U}_X\bm{S}_X\bm{V}_X^T$.
Equality holds in equation~(\ref{the_1a}) 
when $\bm{A}=\bm{U}_X\bm{S}_X^{1/2}$ and $\bm{B}=\bm{S}_X^{1/2}\bm{V}_X^T$;
in equation~(\ref{the_1b}), 
when $\bm{A}=\alpha^{1/3}\bm{U}_X\bm{S}_X^{2/3}$ and $\bm{B}=\alpha^{-1/3}\bm{S}_X^{1/3}\bm{V}_X^T$.

Motivated by this theorem, we define the following factor group-sparse regularizers (FGSR):
\begin{align}
\textup{FGSR}_{1/2}(\bm{X})&:= \frac{1}{2} \min_{\bm{A}\bm{B}=\bm{X}}\ \Vert \bm{A}\Vert_{2,1}+\Vert \bm{B}^T\Vert_{2,1}. \label{Eq.FRM_min_nnzc_12} \\
\textup{FGSR}_{2/3}(\bm{X})&:= \frac{2}{3 \alpha^{1/3}} \min_{\bm{A}\bm{B}=\bm{X}}\ \Vert \bm{A}\Vert_{2,1}+\dfrac{\alpha}{2}\Vert \bm{B}\Vert_F^2, \label{Eq.FRM_min_nnzc_2}
\end{align}
where $\Vert \bm{A}\Vert_{2,1}:=\sum_{j=1}^d\Vert \bm{a}_{j}\Vert$.
Theorem \ref{the_1} shows that $\textup{FGSR}_{2/3}$ has the same value regardless of the choice of $\alpha$,
which justifies the definition.
As a corollary of Theorem \ref{the_1}, we see
\[
\textup{FGSR}_{1/2}(\bm{X}) = \sum_{j=1}^r\sigma_j^{1/2}(\bm{X}) = \Vert \bm{X}\Vert_{\textup{S}_{1/2}}^{1/2},
\qquad
\textup{FGSR}_{2/3}(\bm{X}) = \sum_{j=1}^r\sigma_j^{2/3}(\bm{X}) = \Vert \bm{X}\Vert_{\textup{S}_{2/3}}^{2/3}.
\]

To solve optimization problems involving these surrogates for the rank,
we can use the definition of the FGSR and optimize over the factors $\bm{A}$ and $\bm{B}$.
It is easier to minimize $\textup{FGSR}_{2/3}(\bm{X})$
than to minimize $\textup{FGSR}_{1/2}(\bm{X})$ because the latter has two nonsmooth terms.

As surrogates for the rank function, $\textup{FGSR}_{2/3}$ and $\textup{FGSR}_{1/2}$ have the following advantages:
\begin{itemize}[leftmargin=20pt,itemsep=0pt]
\item \textbf{Tighter rank approximation.}
Compared to the nuclear norm, the spectral quantities in Theorem \ref{the_1} are
tighter approximations to the rank of $\bm{X}$.
\item \textbf{Robust to rank initialization.}
The iterative algorithms we propose in Sections \ref{sec_LRMC} and \ref{sec_RPCA}
to minimize $\textup{FGSR}_{2/3}$ and $\textup{FGSR}_{1/2}$
quickly force some of the columns of $\bm{A}$ and $\bm{B}^T$ to zero, where they remain.
Hence the number of nonzero columns 
is reduced dynamically, and converges to $r$ quickly in experiments:
these methods are \emph{rank-revealing}.
In constrast, iterative methods to minimize the F-nuclear norm or max norm
never produce an exactly-rank-$r$ iterate after a finite number of iterations.
\item \textbf{Low computational cost.}
Most optimization methods for solving problems with the Schatten-$p$ norm
perform SVD on $\bm{X}$ at every iteration,
with time complexity of $O(m^2n)$ (supposing $m\leq n$)
\cite{MC_ShattenP_AAAI125165,Liu2014218}.
In contrast, the natural algorithm to minimize $\textup{FGSR}_{2/3}$ and $\textup{FGSR}_{1/2}$
does not use the SVD, as the regularizers are simple (not spectral) functions of the factors.
The main computational cost is to form $\bm{A}\bm{B}$, which has a time complexity of $O(d'mn)$ when the
iterates $\bm{A}$ and $\bm{B}$ have $d'$ nonzero columns.
The complexity of LRMC can be as low as $O(d'\textup{card}(\Omega))$.
\end{itemize}

\section{Toward exact rank minimization}
In the previous section, we developed a factored representation for $\Vert \bm{X}\Vert_{\textup{S}_{p}}^{p}$
when $p=\tfrac{2}{3}$ or $\tfrac{1}{2}$.
This section develops a similar representation for $\Vert \bm{X}\Vert_{\textup{S}_{p}}^{p}$ with
arbitrarily small $p$.
\begin{theorem}\label{the_2}

Fix $\alpha > 0$, and choose $q\in\lbrace 1,\tfrac{1}{2},\tfrac{1}{4},\cdots\rbrace$.
For any matrix $\bm{X}\in\mathbb{R}^{m\times n}$ with $\textup{rank}(\bm{X})=r\leq d\leq\min(m,n)$, we have
\begin{align}
\min\limits_{\bm{X} = \sum_{j=1}^d\bm{a}_j\bm{b}_j^T}\sum_{j=1}^d\dfrac{1}{q}\Vert \bm{a}_{j}\Vert^q+\alpha\Vert \bm{b}_{j}\Vert
=& (1+1/q)\alpha^{q/(q+1)}\sum_{j=1}^r\sigma_j^{q/(q+1)}(\bm{X}) \label{the_2a} , \\
\min\limits_{\bm{X} = \sum_{j=1}^d\bm{a}_j\bm{b}_j^T}\sum_{j=1}^d\dfrac{1}{q}\Vert \bm{a}_{j}\Vert^q+\dfrac{\alpha}{2}\Vert \bm{b}_{j}\Vert^2
=& (1/2+1/q)\alpha^{q/(q+2)}\sum_{j=1}^r\sigma_j^{2q/(2+q)}(\bm{X}). \label{the_2b}
\end{align}

\end{theorem}
By choosing an appropriate $q$,
these representations give arbitrarily tight approximations to the rank, since
$\Vert \bm{X}\Vert_{\textup{S}_{p}}^{p} \to \textup{rank}(\bm{X})$ as $p \to 0$.
For example, use (\ref{the_2b}) in Theorem \ref{the_2} when $q=\tfrac{1}{4}$ to see
\begin{equation}
\min\limits_{\sum_{j=1}^d\bm{a}_j\bm{b}_j^T=\bm{X}}\sum_{j=1}^d\dfrac{1}{1/4}\Vert \bm{a}_{j}\Vert^{1/4}+\dfrac{\alpha}{2}\Vert \bm{b}_{j}\Vert^2= 4.5\alpha^{1/9}\sum_{i=1}^d\sigma_i^{2/9}(\bm{X})=4.5\alpha^{1/9}\Vert \bm{X}\Vert_{\textup{S}_{2/9}}^{2/9}.
\end{equation}

Equality holds in equation~(\ref{the_2a}) 
when $\bm{A}=\alpha^{{1}/{(q+1)}}\bm{U}_X\bm{S}_X^{{1}/{(q+1)}}$ and $\bm{B}=\alpha^{{-1}/{(q+1)}}\bm{S}_X^{{q}/{(q+1)}}\bm{V}_X^T$;
in equation~(\ref{the_2b}), 
when
$\bm{A}=\alpha^{{1}/{(q+2)}}\bm{U}_X\bm{S}_X^{{2}/{(q+2)}}$ and $\bm{B}=\alpha^{{-1}/{(q+2)}}\bm{S}_X^{{q}/{(q+2)}}\bm{V}_X^T$.

\section{Application to low-rank matrix completion}\label{sec_LRMC}
As an application, we model noiseless matrix completion using $\textup{FGSR}$
as a surrogate for the rank:
\begin{equation}\label{Eq.FRM_LRMC_noisy_0}
\mathop{\textup{minimize}}_{\bm{X}}\  \textup{FGSR}(\bm{X}) ,\quad
\textup{subject to}\  P_{\Omega}(\bm{X})=P_{\Omega}(\bm{M}).
\end{equation}
Take $ \textup{FGSR}_{2/3}$ as an example. We rewrite (\ref{Eq.FRM_LRMC_noisy_0}) as
\begin{equation}\label{Eq.FRM_LRMC_noisy}
\mathop{\textup{minimize}}_{\bm{X},\bm{A},\bm{B}}\  \Vert \bm{A}\Vert_{2,1}+\dfrac{\alpha}{2}\Vert \bm{B}\Vert_F^2,\quad
\textup{subject to}\ \bm{X}=\bm{A}\bm{B},\ P_{\Omega}(\bm{X})=P_{\Omega}(\bm{M}).
\end{equation}
This problem is separable in the three blocks of unknowns $\bm{X}$, $\bm{A}$, and $\bm{B}$.
We propose to use the Alternating Direction Method of Multipliers (ADMM) \cite{wang2015global,liu2017linearized,gao2018admm} with linearization to solve this problem,
as the ADMM subproblem for $\bm{A}$ has no closed-form solution. Details are in the supplement.

Now consider an application to noisy matrix completion.
Suppose we observe $P_{\Omega}(\bm{M}_e)$ with $\bm{M}_e=\bm{M}+\bm{E}$, where $\bm{E}$ represents measurement noise.
Model the problem using 
$\textup{FGSR}_{2/3}$ as
\begin{equation}\label{Eq.FRM_LRMC_noisy_1}
\mathop{\textup{minimize}}_{\bm{A},\bm{B}}\ \Vert \bm{A}\Vert_{2,1}+\dfrac{\alpha}{2}\Vert \bm{B}\Vert_F^2+\dfrac{\beta}{2}\Vert P_{\Omega}(\bm{M_e}-\bm{A}\bm{B})\Vert_F^2.
\end{equation}
We can still solve the problem via linearized ADMM.
However, proximal alternating linearized minimization (PALM) \cite{bolte2014proximal,FAN2018SFMC} gives a more  efficient method. Details are in the supplement.

Motivated by Theorem \ref{the_2}, we can also
model noisy matrix completion with a sharper rank surrogate:
\begin{equation}\label{Eq.GP1}
\mathop\textup{minimize}_{\bm{A},\bm{B}}\dfrac{1}{2}\Vert\mathcal{P}_{\Omega}(\bm{M}_e-\bm{A}\bm{B})\Vert_F^2+\gamma\Big(\dfrac{1}{q}\Vert \bm{A}\Vert_{2,q}^{q}+\dfrac{\alpha}{2}\Vert \bm{B}^T\Vert_F^2\Big),
\end{equation}
where $q\in\lbrace 1,\tfrac{1}{2},\tfrac{1}{4},\cdots\rbrace$ and $\Vert \bm{A}\Vert_{2,q}:=\Big(\sum_{j=1}^d\Vert\bm{a}_{j}\Vert^q\Big)^{1/q}$.
When $q < 1$, we suggest solving the problem~(\ref{Eq.GP1}) using PALM coupled with iteratively reweighted minimization \cite{mohan2012iterative}. According to the number of degrees of freedom of low-rank matrix, we suggest $d=\vert\Omega\vert/(m+n)$ in practical applications. 

\section{Generalization error bound for LRMC}\label{sec_gb}
Above, we proposed a method to solve LRMC problems
using a FGSR as a rank surrogate.
Here, we develop an upper bound on the error of the resulting estimator
using a new generalization bound for LRMC with a Schatten-$p$ norm constraint.
Similar bounds are available 
for LRMC using the nuclear norm \cite{srebro2005rank} and max norm \cite{foygel2011concentration}.


Consider the following observation model. A matrix $\bm{M}$ is corrupted
with iid $\mathcal{N}(0,\epsilon^2)$ noise $\bm{E}$ to form $\bm{M}_e = \bm{M} + \bm{E}$.
Suppose each entry of $\bm{M}_e$ is observed independently with probability $\rho$ and the number of observed entries is $\vert\Omega\vert$, where $\mathbb{E} |\Omega|=\rho mn$.

Choose $q\in\lbrace 1,\tfrac{1}{2},\tfrac{1}{4},\cdots\rbrace$ and $p=\tfrac{2q}{2+q}$.
For any $\gamma>0$, consider a solution $(\bm{A},\bm{B})$ to (\ref{Eq.GP1}).
Let $\Vert\bm{A}\bm{B}\Vert_{\textup{S}_p}^p=R_p$.
Then use Theorem \ref{the_2} to see that the following problem has the same solution,
\begin{equation}\label{Eq.GP0}
\mathop\textup{minimize}_{\Vert\bm{X}\Vert_{\textup{S}_p}^p\leq R_p,\textup{rank}(\bm{X})\leq d}\Vert\mathcal{P}_{\Omega}(\bm{M}_e-\bm{X})\Vert_F^2.
\end{equation}
Therefore, we may solve (\ref{Eq.GP1}) using the methods described above to find a solution to (\ref{Eq.GP0}) efficiently.
In this section, we provide generalization error bounds for the solution $\hat{\bm{M}}$
of (\ref{Eq.GP0}).

\subsection{Bound with optimal solution}

Without loss of generality, we may assume $\Vert\bm{M}\Vert_\infty\leq \varsigma/\sqrt{mn}$ for some constant $\varsigma$. Hence it is reasonable to assume that $\epsilon=\epsilon_0/\sqrt{mn}$ for some constant $\epsilon_0$. The following theorem provides a generalization error bound for the solution of (\ref{Eq.GP0}).
\begin{theorem}\label{the_4}
Suppose $\Vert\bm{M}\Vert_{\textup{S}_p}^p\leq R_p$, $\hat{\bm{M}}$ is the optimal solution of (\ref{Eq.GP0}), and $\vert\Omega\vert\geq \tfrac{32}{3}n\log^2n$. Denote $\zeta:=\max\lbrace\Vert\bm{M}\Vert_\infty,\Vert\hat{\bm{M}}\Vert_\infty\rbrace$. Then there exist numerical constants $c_1$ and $c_2$ such that the following inequality holds with probability at least $1-5n^{-2}$
\begin{equation}\label{the_4_ineq}
\Vert\bm{M}-\hat{\bm{M}}\Vert_F^2\leq \max\left\lbrace c_1\zeta^2\dfrac{n\log n}{\vert\Omega\vert},(5.5+\sqrt{10})R_p\Bigg((4\sqrt{3}\epsilon_0+c_2\zeta)^2\dfrac{n\log n}{\vert\Omega\vert} \Bigg)^{1-p/2} \right\rbrace.
\end{equation}
\end{theorem}
\vspace{-10pt}
When $\vert \Omega\vert$ is sufficiently large, we see that the second term in the brace of (\ref{the_4_ineq}) is the dominant term, which decreases as $p$ decreases.
A more complicated but more informative bound can be found in the supplement (inequality (24)).
In sum, Theorem \ref{the_4} shows it is possible to reduce the matrix completion error
by using a smaller $p$ in (\ref{Eq.GP0}) or a smaller $q$ in (\ref{Eq.GP1}).

\subsection{Bound with arbitrary \textit{A} and \textit{B}} %
Since (\ref{Eq.GP1}) and (\ref{Eq.GP0}) are nonconvex problems,
it is difficult to guarantee that an optimization method has found a globally optimal solution.
The following theorem provides a generalization bound
for any feasible point $(\hat{\bm{A}}, \hat{\bm{B}})$ of (\ref{Eq.GP1}):
\begin{theorem}\label{the_3}
Suppose $\bm{M}_e = \bm{M} + \bm{E}$.
For any $\hat{\bm{A}}$ and $\hat{\bm{B}})$,
let $\hat{\bm{M}}=\hat{\bm{A}}\hat{\bm{B}}$ and $d$ be the number of nonzero columns of $\hat{\bm{A}}$.
Define $\zeta:=\max\lbrace\Vert\bm{M}\Vert_\infty,\Vert\hat{\bm{M}}\Vert_\infty\rbrace$.
Then there exists a numerical constant $C_0$, such that with probability at least $1-2\exp(-n)$, the following equality holds:
\begin{equation*}
\dfrac{\Vert \bm{M}-\hat{\bm{M}}\Vert_F}{\sqrt{mn}}
\leq \dfrac{\Vert \mathcal{P}_{\Omega}(\bm{M}_e-\hat{\bm{M}})\Vert_F}{\sqrt{\vert\Omega\vert}}+\dfrac{\Vert \bm{E}\Vert_F}{\sqrt{mn}}+C_0\zeta\Big(\dfrac{nd\log{n}}{\vert\Omega\vert}\Big)^{1/4}.
\end{equation*}
\end{theorem}
\vspace{-10pt}
Theorem \ref{the_3} indicates that if the training error $ \Vert \mathcal{P}_{\Omega}(\bm{M}_e-\hat{\bm{A}}\hat{\bm{B}})\Vert_F $ and the number $ d $ of nonzero columns of $\hat{\bm{A}}$ are small, the matrix completion error is small.
In particular, if $\bm{E}=\bm{0}$ and $\mathcal{P}_{\Omega}(\bm{M}_e-\hat{\bm{A}}\hat{\bm{B}})=\bm{0}$,
the matrix completion error is upper-bounded by $C_0\zeta\big(\tfrac{nd\log{n}}{\vert\Omega\vert}\big)^{1/4}$.
We hope that a smaller $q$ in (\ref{Eq.GP1}) can lead to smaller training error and $d$ such that the upper bound of matrix completion error is smaller.
Indeed, in our experiments, we find that smaller $q$ often leads to smaller matrix completion error,
but the improvement saturates quickly as $q$ decreases.
We find $q=1$ or $\tfrac{1}{2}$ (corresponding to a Schatten-$p$ norm with $p=\tfrac{2}{3}$ or $\tfrac{2}{5}$)
are enough to provide high matrix completion accuracy and outperform max norm and nuclear norm.

%

\section{Application to robust PCA}\label{sec_RPCA}
Suppose a fraction of entries in a matrix are corrupted in random locations.
Formally, we observe
\begin{equation}\label{RPCA_p0}
\bm{M}_e=\bm{M}+\bm{E},
\end{equation}
where $\bm{M}$ is a low-rank matrix and $\bm{E}$ is the sparse corruption matrix whose nonzero entries may be arbitrary.
The robust principal component analysis (RPCA) asks to recover $\bm{M}$ from $\bm{M}_e$;
a by-now classic approach uses nuclear norm minimization \cite{RPCA}.
We propose to use FGSR instead, and solve
\begin{equation}\label{Eq.FRM_RPCA}
\mathop{\textup{minimize}}_{\bm{A},\bm{B},\bm{E}}\  \dfrac{1}{q}\Vert \bm{A}\Vert_{2,q}^q+\dfrac{\alpha}{2}\Vert \bm{B}\Vert_F^2+\lambda\Vert \bm{E}\Vert_1,\quad
\textup{subject to}\ \bm{M}_e=\bm{A}\bm{B}+\bm{E},
\end{equation}
where $q\in\lbrace 1,\tfrac{1}{2},\tfrac{1}{4},\cdots\rbrace$. An optimization algorithm is detailed in the supplement.

\section{Numerical results}
\subsection{Matrix completion}
\paragraph{Baseline methods}
We compare the FGSR regularizers with the nuclear norm, truncated nuclear norm \cite{MC_TNNR_PAMI2013}, weighted nuclear norm \cite{gu2014weighted}, F-nuclear norm, max norm \cite{lee2010practical}, Riemannian pursuit \cite{tan2014riemannian}, Schatten-$p$ norm, Bi-nuclear norm \cite{shang2016tractable}, and F$^2$+nuclear norm \cite{shang2017bilinear}.
We choose the parameters of all methods to ensure they perform as well as possible.
Details about the optimizations, parameters, evaluation metrics are in the supplement.
All experiments present the average of ten trials.

\paragraph{Noiseless synthetic data}
We generate random matrices of size $500\times 500$ and rank $50$.
More details about the experiment are in the supplement.
In Figure \ref{Fig_MC_clean_curve}(a),
the factored methods all use factors of size $d=1.5r$.
We see the Schatten-$p$ norm ($p=\tfrac{2}{3}$,$\tfrac{1}{2}$,$\tfrac{1}{4})$, Bi-nuclear norm, F$^2$+nuclear norm, FGSR$_{2/3}$, and FGSR$_{1/2}$
have similar performances and outperform other methods when the \emph{missing rate} (proportion of unobserved entries) is high.
In particular, the F-nuclear norm outperforms the nuclear norm because the bound $d$ on the rank is binding.
In Figure \ref{Fig_MC_clean_curve}(b) and (c), in which the missing rates are high,
the max norm and F-nuclear norm are sensitive to the initial rank $d$,
while the F$^2$+nuclear norm, Bi-nuclear norm, FGSR$_{2/3}$, and FGSR$_{1/2}$  always have nearly zero recovery error.
Interestingly, the max norm and F-nuclear norm are robust to the initial rank
when the missing rate is much lower than 0.6 in this experiment.
In Figure \ref{Fig_MC_clean_curve}(d), we compare the computational time in the case of missing rate$=0.7$,
in which, for fair comparison, the optimization algorithms of all methods were stopped
when the relative change of the recovered matrix was less than $10^{-5}$ or the number of iterations reached 1000.
The computational cost of nuclear norm, truncated nuclear norm, weighted nuclear norm, and Schatten-$\tfrac{1}{2}$ norm are especially large,
as they require computing the SVD in every iteration.
The computational costs of max norm, F-nuclear norm, F$^2$+nuclear norm, and Bi-nuclear norm  increase quickly as the initial rank $d$ increases.
In contrast, our FGSR$_{2/3}$ and FGSR$_{1/2}$  are very efficient even when the initial rank is large, because they are SVD-free and able to reduce the size of the factors in the progress of optimization.
While Riemannian pursuit is a bit faster than FGSR, FGSR has lower error.
Note that the Riemannian pursuit code mixes C and MATLAB, while all other methods are written in pure MATLAB,
explaining (part of) its more nimble performance.

\begin{figure*}[h!]
\centering
\includegraphics[width=11cm,trim={50 0 170 0}, clip]{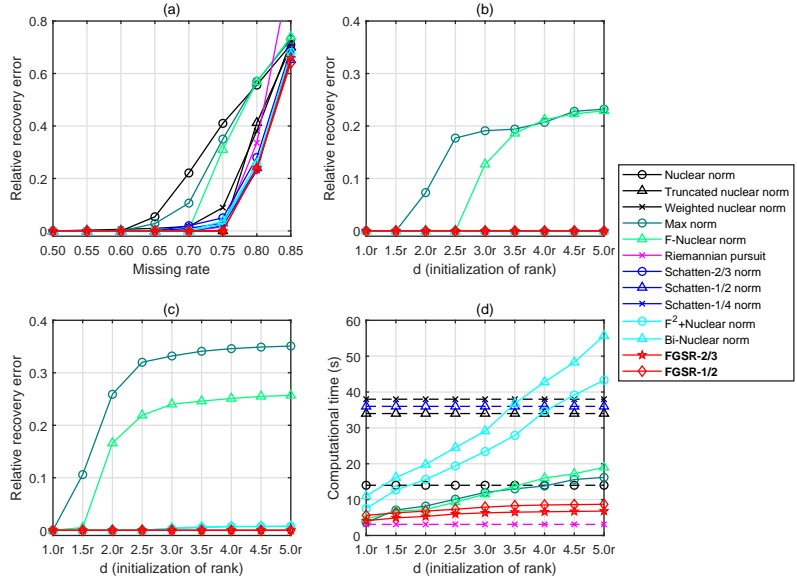}
\abovecaptionskip -2pt
\caption{Matrix completion on noiseless synthetic data ($r=50$): (a) the effect of missing rate on recovery error; (b)(c) the effect of rank initialization on recovery error ($\textup{missing rate}=0.6$ or $0.7$); (d) the effect of rank initialization on computational cost ($\textup{missing rate}=0.7$).}
\label{Fig_MC_clean_curve}
\end{figure*}

\begin{figure*}[h!]
\centering
\includegraphics[width=9cm,trim={20 0 20 0}, clip]{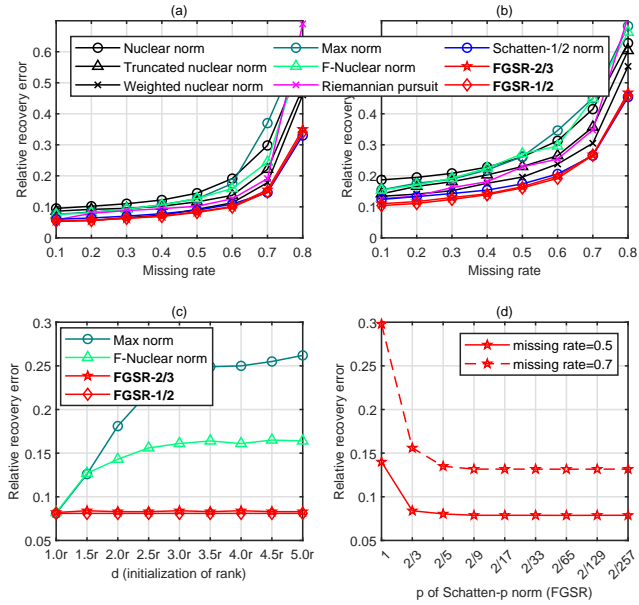}
\abovecaptionskip -2pt
\caption{Matrix completion on noisy synthetic data: (a)(b) recovery error when $\textup{SNR}=10$ or $5$; (c) the effect of rank initialization on recovery error ($\textup{SNR}=10$, $\textup{missing rate}=0.5$); (d) the effect of $p$ in Schatten-$p$ norm (using FGSR when $p<1$).}
\label{MC_syn_noisy}
\end{figure*}

\paragraph{Noisy synthetic data}
We simulate a noisy matrix completion problem by adding Gaussian noise to low-rank random matrices.
We omit F$^2$+nuclear norm and Bi-nuclear norm from these results
because they are less efficient that FGSR$_{2/3}$ and FGSR$_{1/2}$ but perform similarly on recovery error.
The recovery errors for different missing rate are reported in Figure \ref{MC_syn_noisy} (a) and (b) for $\textup{SNR}=10$ and $\textup{SNR}=5$ (SNR$:=\Vert\bm{M}\Vert_F/\Vert\bm{E}\Vert_F$) respectively.
The max norm outperforms the nuclear norm when the missing rate is low.
The recovery errors of Schatten-$\tfrac{1}{2}$ norm, FGSR$_{2/3}$, and FGSR$_{1/2}$ are much lower than those of others.
Figure \ref{MC_syn_noisy}(c) demonstrates that our FGSR$_{2/3}$ and FGSR$_{1/2}$ are robust to the initial rank,
while max norm and F-nuclear norm degrade as the initial rank increases.
In Figure \ref{MC_syn_noisy}(d), we see decreasing $p$ from $1$ to $2/9$ reduces the recovery error significantly,
but the recovery error stabilizes for smaller $p$. This result is consistent with Theorem \ref{the_4}.

\begin{figure*}[h!]
\centering
\includegraphics[width=9cm,trim={20 0 20 0}, clip]{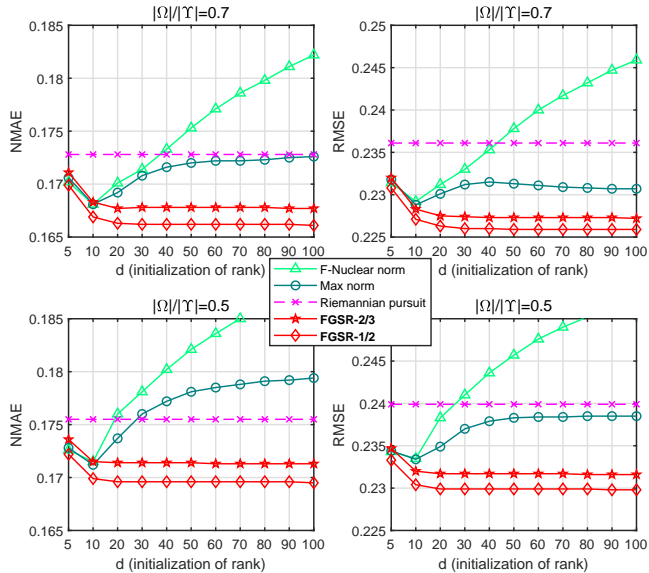}
\abovecaptionskip -2pt
\caption{NMAE and RMSE  on Movielens-1M data ($\Upsilon$: known entries; $\Omega$: sampled entries from $\Upsilon$)}
\label{MC_mv1M}
\end{figure*}

\paragraph{Real data}
We consider the MovieLens-1M dataset \cite{Web_MovieLens1M}, which consists of 1 million ratings ($1$ to $5$)
for 3900 movies by 6040 users. The movies rated by less than 5 users are deleted in this study because the corresponding ratings may never be recovered when the matrix rank is higher than 5.
We randomly sample $70\%$ or $50\%$ of the known ratings of each user and perform matrix completion.
The normalized mean absolute error (NMAE) \cite{toh2010accelerated,shamir2014matrix} and normalized root-mean-squared-error (RMSE) \cite{shamir2014matrix} are reported in Figure \ref{MC_mv1M}, in which each value is the average of ten repeated trials and the standard deviation is less than $0.0003$. Although Riemannian pursuit can adaptively determine the rank, its performance is not satisfactory.
As the initial rank increases, the NMAE and RMSE of max norm and F-nuclear norm increase.
In contrast, FGSR$_{2/3}$ and FGSR$_{1/2}$ have consistent low NMAE and RMSE. Moreover, FGSR$_{1/2}$ outperforms FGSR$_{2/3}$.

\subsection{Robust PCA}
\begin{figure*}[h!]
\centering
\includegraphics[width=9cm,trim={20 0 20 0}, clip]{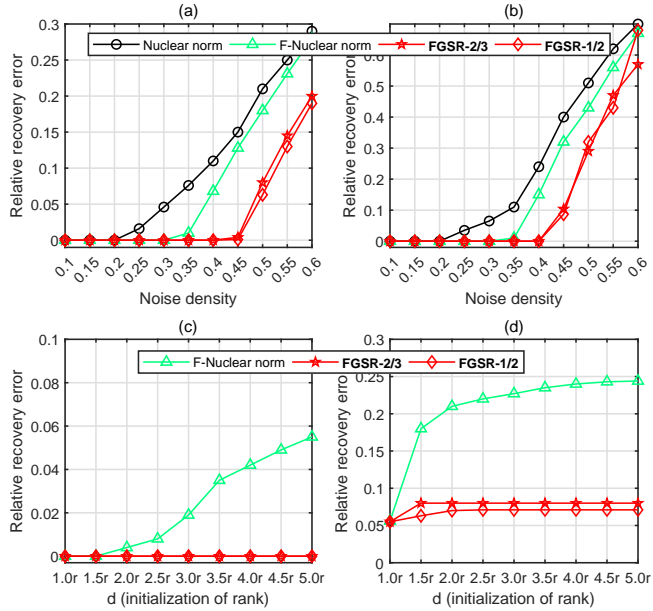}
\abovecaptionskip -2pt
\caption{RPCA on synthetic data: (a)(b) recovery error when $\textup{SNR}_c=1$ or $0.2$; (c)(d) the effect of rank initialization on recovery error ($\textup{SNR}_c=1$, $\textup{noise density}=0.3$ or $0.5$).}
\label{RPCA_syn_curve}
\end{figure*}

We simulate a corrupted matrix as $\bm{M}_e=\bm{M}+\bm{E}$,
where $\bm{M}$ is a random matrix of size $500\times 500$ with rank $50$
and $\bm{E}$ is a sparse matrix whose nonzero entries are $\mathcal{N}(0,\epsilon^2)$.
Define the signal-noise-ratio $\textup{SNR}_c:=\sigma/\epsilon$,
where $\sigma$ denotes the standard deviation of the entries of $\bm{M}$.
Figure \ref{RPCA_syn_curve}(a) and (b) show the recovery errors
for different noise densities (proportion of nonzero entries of $\bm{E}$).
When the noise density is high, FGSR$_{2/3}$ and FGSR$_{1/2}$ outperform nuclear norm and F-nuclear norm.
Figure \ref{RPCA_syn_curve}(c) and (d) shows
again that unlide the F-nuclear norm, FGSR$_{2/3}$ and  FGSR$_{1/2}$ are not sensitive to the initial rank,
and that FGSR$_{1/2}$ outperforms FGSR$_{2/3}$ slightly when the noise density is high.
More results, including for image denoising, appear in the supplement.

\vspace{-5pt}

\vspace{-5pt}

\section{Conclusion}
This paper proposed a class of nonconvex surrogates for matrix rank
that we call Factor Group-Sparse Regularizers (FGSRs).
These FGSRs give a factored formulation for certain Schatten-$p$ norms
with arbitrarily small $p$.
These FGSRs are tighter surrogates for the rank than the nuclear norm,
can be optimized without the SVD,
and perform well in denoising and completion tasks regardless of the initial choice of rank.
In addition, we provide generalization error bounds for LRMC using the FGSR
(or, more generally, any Schatten-$p$ norm) as a regularizer.
Our experimental results demonstrate the proposed methods\footnote{The MATLAB codes of the proposed methods are available at \emph{https://github.com/udellgroup/Codes-of-FGSR-for-effecient-low-rank-matrix-recovery}} achieve state-of-the-art performances in LRMC and RPCA.

These experiments provide compelling evidence that PALM and ADMM may often
(perhaps always) converge to the global optimum of these problems.
A full convergence theory is an interesting problem for future work.
A proof of global convergence would reveal the required sample complexity for
LRMC and RPCA with FGSR as a computationally tractable rank proxy.

\newpage




\bibliographystyle{unsrt}
\small
\bibliography{Ref_FRM}

\begin{thebibliography}{10}

\bibitem{udell2019big}
Madeleine Udell and Alex Townsend.
\newblock Why are big data matrices approximately low rank?
\newblock {\em SIAM Journal on Mathematics of Data Science}, 1(1):144--160,
  2019.

\bibitem{CandesRecht2009}
Emmanuel~J. Cand\`{e}s and Benjamin Recht.
\newblock Exact matrix completion via convex optimization.
\newblock {\em Foundations of Computational Mathematics}, 9(6):717--772, 2009.

\bibitem{toh2010accelerated}
Kim-Chuan Toh and Sangwoon Yun.
\newblock An accelerated proximal gradient algorithm for nuclear norm
  regularized linear least squares problems.
\newblock {\em Pacific Journal of optimization}, 6(615-640):15, 2010.

\bibitem{MC_Recht2011:SAM}
Benjamin Recht.
\newblock A simpler approach to matrix completion.
\newblock {\em J. Mach. Learn. Res.}, 12:3413--3430, December 2011.

\bibitem{foygel2011concentration}
Rina Foygel and Nathan Srebro.
\newblock Concentration-based guarantees for low-rank matrix reconstruction.
\newblock In {\em Proceedings of the 24th Annual Conference on Learning
  Theory}, pages 315--340, 2011.

\bibitem{hardt2014understanding}
Moritz Hardt.
\newblock Understanding alternating minimization for matrix completion.
\newblock In {\em 2014 IEEE 55th Annual Symposium on Foundations of Computer
  Science (FOCS),}, pages 651--660. IEEE, 2014.

\bibitem{MC_icml2014c1_chenc14}
Yudong Chen, Srinadh Bhojanapalli, Sujay Sanghavi, and Rachel Ward.
\newblock Coherent matrix completion.
\newblock In {\em Proceedings of the 31st International Conference on Machine
  Learning (ICML-14)}, pages 674--682. JMLR Workshop and Conference
  Proceedings, 2014.

\bibitem{shamir2014matrix}
Ohad Shamir and Shai Shalev-Shwartz.
\newblock Matrix completion with the trace norm: learning, bounding, and
  transducing.
\newblock {\em The Journal of Machine Learning Research}, 15(1):3401--3423,
  2014.

\bibitem{sun2016guaranteed}
Ruoyu Sun and Zhi-Quan Luo.
\newblock Guaranteed matrix completion via non-convex factorization.
\newblock {\em IEEE Transactions on Information Theory}, 62(11):6535--6579,
  2016.

\bibitem{Fan_2019_CVPR}
Jicong Fan and Madeleine Udell.
\newblock Online high rank matrix completion.
\newblock In {\em The IEEE Conference on Computer Vision and Pattern
  Recognition (CVPR)}, June 2019.

\bibitem{NIPS2010_3932}
Andrew Goldberg, Ben Recht, Junming Xu, Robert Nowak, and Xiaojin Zhu.
\newblock Transduction with matrix completion: Three birds with one stone.
\newblock In {\em Advances in Neural Information Processing Systems 23}, pages
  757--765. Curran Associates, Inc., 2010.

\bibitem{udell2016generalized}
Madeleine Udell, Corinne Horn, Reza Zadeh, Stephen Boyd, et~al.
\newblock Generalized low rank models.
\newblock {\em Foundations and Trends{\textregistered} in Machine Learning},
  9(1):1--118, 2016.

\bibitem{RPCA}
Emmanuel~J. Cand\`{e}s, Xiaodong Li, Yi~Ma, and John Wright.
\newblock Robust principal component analysis?
\newblock {\em J. ACM}, 58(3):1--37, 2011.

\bibitem{feng2013online}
Jiashi Feng, Huan Xu, and Shuicheng Yan.
\newblock Online robust {PCA} via stochastic optimization.
\newblock In {\em Advances in Neural Information Processing Systems}, pages
  404--412, 2013.

\bibitem{zhao2014robust}
Qian Zhao, Deyu Meng, Zongben Xu, Wangmeng Zuo, and Lei Zhang.
\newblock Robust principal component analysis with complex noise.
\newblock In {\em International Conference on Machine Learning}, pages 55--63,
  2014.

\bibitem{pimentel2017random}
Daniel Pimentel-Alarc{\'o}n and Robert Nowak.
\newblock Random consensus robust {PCA}.
\newblock In {\em Artificial Intelligence and Statistics}, pages 344--352,
  2017.

\bibitem{rkpca_fan}
J.~{Fan} and T.~W.~S. {Chow}.
\newblock Exactly robust kernel principal component analysis.
\newblock {\em IEEE Transactions on Neural Networks and Learning Systems},
  pages 1--13, 2019.

\bibitem{8425659}
T.~{Bouwmans}, S.~{Javed}, H.~{Zhang}, Z.~{Lin}, and R.~{Otazo}.
\newblock On the applications of robust {PCA} in image and video processing.
\newblock {\em Proceedings of the IEEE}, 106(8):1427--1457, Aug 2018.

\bibitem{MC_TNNR_PAMI2013}
Yao Hu, Debing Zhang, Jieping Ye, Xuelong Li, and Xiaofei He.
\newblock Fast and accurate matrix completion via truncated nuclear norm
  regularization.
\newblock {\em IEEE Transactions on Pattern Analysis and Machine Intelligence},
  35(9):2117--2130, 2013.

\bibitem{gu2014weighted}
Shuhang Gu, Lei Zhang, Wangmeng Zuo, and Xiangchu Feng.
\newblock Weighted nuclear norm minimization with application to image
  denoising.
\newblock In {\em Proceedings of the IEEE conference on computer vision and
  pattern recognition}, pages 2862--2869, 2014.

\bibitem{MC_ShattenP_AAAI125165}
Feiping Nie, Heng Huang, and Chris Ding.
\newblock Low-rank matrix recovery via efficient {Schatten} p-norm
  minimization.
\newblock In {\em Proceedings of the Twenty-Sixth AAAI Conference on Artificial
  Intelligence}, AAAI'12, pages 655--661. AAAI Press, 2012.

\bibitem{Liu2014218}
Lu~Liu, Wei Huang, and Di-Rong Chen.
\newblock Exact minimum rank approximation via {Schatten} p-norm minimization.
\newblock {\em Journal of Computational and Applied Mathematics}, 267:218 --
  227, 2014.

\bibitem{pmlr-v70-ongie17a}
Greg Ongie, Rebecca Willett, Robert~D. Nowak, and Laura Balzano.
\newblock Algebraic variety models for high-rank matrix completion.
\newblock In {\em Proceedings of the 34th International Conference on Machine
  Learning}, pages 2691--2700, Sydney, Australia, 06--11 Aug 2017. PMLR.

\bibitem{mohan2012iterative}
Karthik Mohan and Maryam Fazel.
\newblock Iterative reweighted algorithms for matrix rank minimization.
\newblock {\em Journal of Machine Learning Research}, 13(Nov):3441--3473, 2012.

\bibitem{srebro2005maximum}
Nathan Srebro, Jason Rennie, and Tommi~S. Jaakkola.
\newblock Maximum-margin matrix factorization.
\newblock In {\em Advances in neural information processing systems}, pages
  1329--1336, 2005.

\bibitem{rennie2005fast}
Jasson~DM Rennie and Nathan Srebro.
\newblock Fast maximum margin matrix factorization for collaborative
  prediction.
\newblock In {\em Proceedings of the 22nd international conference on Machine
  learning}, pages 713--719. ACM, 2005.

\bibitem{NIPS2010_4102}
Nathan Srebro and Ruslan~R Salakhutdinov.
\newblock Collaborative filtering in a non-uniform world: Learning with the
  weighted trace norm.
\newblock In J.~D. Lafferty, C.~K.~I. Williams, J.~Shawe-Taylor, R.~S. Zemel,
  and A.~Culotta, editors, {\em Advances in Neural Information Processing
  Systems 23}, pages 2056--2064. Curran Associates, Inc., 2010.

\bibitem{MC_MF_Wen2012}
Zaiwen Wen, Wotao Yin, and Yin Zhang.
\newblock Solving a low-rank factorization model for matrix completion by a
  nonlinear successive over-relaxation algorithm.
\newblock {\em Mathematical Programming Computation}, 4(4):333--361, 2012.

\bibitem{tan2014riemannian}
Mingkui Tan, Ivor~W Tsang, Li~Wang, Bart Vandereycken, and Sinno~Jialin Pan.
\newblock Riemannian pursuit for big matrix recovery.
\newblock In {\em International Conference on Machine Learning}, pages
  1539--1547, 2014.

\bibitem{srebro2005rank}
Nathan Srebro and Adi Shraibman.
\newblock Rank, trace-norm and max-norm.
\newblock In {\em International Conference on Computational Learning Theory},
  pages 545--560. Springer, 2005.

\bibitem{lee2010practical}
Jason~D. Lee, Ben Recht, Nathan Srebro, Joel Tropp, and Ruslan~R.
  Salakhutdinov.
\newblock Practical large-scale optimization for max-norm regularization.
\newblock In {\em Advances in Neural Information Processing Systems}, pages
  1297--1305, 2010.

\bibitem{gunasekar2017implicit}
Suriya Gunasekar, Blake~E Woodworth, Srinadh Bhojanapalli, Behnam Neyshabur,
  and Nati Srebro.
\newblock Implicit regularization in matrix factorization.
\newblock In {\em Advances in Neural Information Processing Systems}, pages
  6151--6159, 2017.

\bibitem{shang2016tractable}
Fanhua Shang, Yuanyuan Liu, and James Cheng.
\newblock Tractable and scalable {Schatten} quasi-norm approximations for rank
  minimization.
\newblock In {\em Artificial Intelligence and Statistics}, pages 620--629,
  2016.

\bibitem{shang2017bilinear}
Fanhua Shang, James Cheng, Yuanyuan Liu, Zhi-Quan Luo, and Zhouchen Lin.
\newblock Bilinear factor matrix norm minimization for robust pca: Algorithms
  and applications.
\newblock {\em IEEE transactions on pattern analysis and machine intelligence},
  40(9):2066--2080, 2017.

\bibitem{wang2015global}
Yu~Wang, Wotao Yin, and Jinshan Zeng.
\newblock Global convergence of admm in nonconvex nonsmooth optimization.
\newblock {\em Journal of Scientific Computing}, pages 1--35, 2015.

\bibitem{liu2017linearized}
Qinghua Liu, Xinyue Shen, and Yuantao Gu.
\newblock Linearized admm for non-convex non-smooth optimization with
  convergence analysis.
\newblock {\em arXiv preprint arXiv:1705.02502}, 2017.

\bibitem{gao2018admm}
Wenbo Gao, Donald Goldfarb, and Frank~E Curtis.
\newblock Admm for multiaffine constrained optimization.
\newblock {\em arXiv preprint arXiv:1802.09592}, 2018.

\bibitem{bolte2014proximal}
J{\'e}r{\^o}me Bolte, Shoham Sabach, and Marc Teboulle.
\newblock Proximal alternating linearized minimization for nonconvex and
  nonsmooth problems.
\newblock {\em Mathematical Programming}, 146(1-2):459--494, 2014.

\bibitem{FAN2018SFMC}
J.~Fan, M.~Zhao, and T.~W.~S. Chow.
\newblock Matrix completion via sparse factorization solved by accelerated
  proximal alternating linearized minimization.
\newblock {\em IEEE Transactions on Big Data}, pages 1--1, 2018.

\bibitem{Web_MovieLens1M}
{MovieLens} dataset.
\newblock {\em https://grouplens.org/datasets/movielens/}.

\end{thebibliography}

\end{document}